%% file: ICRA2024_MRIO.tex
\documentclass[letterpaper, 10 pt, conference]{ieeeconf}  

\IEEEoverridecommandlockouts                              

\usepackage{cite}
\usepackage{amsmath,amssymb,amsfonts}
\usepackage{algorithmic}
\usepackage{graphicx}
\usepackage{textcomp}
\usepackage{xcolor}

\usepackage[font=footnotesize,labelfont=bf]{caption}
\usepackage[font=footnotesize,labelfont=bf]{subcaption}
\usepackage{balance}
\usepackage[group-separator={,}, per-mode=symbol]{siunitx}
\DeclareMathOperator*{\argmin}{argmin}

\makeatletter
\let\NAT@parse\undefined
\makeatother
\usepackage{hyperref}
\usepackage{cleveref}
\hypersetup{
    colorlinks=true,
    linkcolor=blue,
    filecolor=magenta,      
    urlcolor=cyan,
    pdfpagemode=FullScreen,
}

\title{\LARGE \bf
Multi-Radar Inertial Odometry for 3D State Estimation \\
using mmWave Imaging Radar
}

\author{Jui-Te Huang, Ruoyang Xu, Akshay Hinduja, and Michael Kaess
\thanks{This work was partially supported by Amazon Lab126.}
\thanks{J. Huang, A. Hinduja and M. Kaess are, R. Xu was, with the Robotics Institute, Carnegie Mellon University, Pittsburgh, PA 15213, USA \tt{\{juiteh, ruoyangx, ahinduja, kaess\}@cs.cmu.edu}}%
}

\begin{document}

\maketitle

\input{mathabbreviations}
\input{00_abstract.tex}

\input{01_introduction.tex}
\input{02_related_work.tex}

\input{04_methodology.tex}
\input{05_experiments.tex}
\input{06_conclusion.tex}

\bibliographystyle{IEEEtran}
\balance
\bibliography{ref}

\end{document}

%% file: mathabbreviations.tex

\newcommand{\vc}[1]{\boldsymbol{#1}}
\newcommand{\adj}[1]{\frac{d J}{d #1}}
\newcommand{\chain}[2]{\adj{#2} = \adj{#1}\frac{d #1}{d #2}}

\newcommand{\Ac}{\mathcal{A}}
\newcommand{\Bc}{\mathcal{B}}
\newcommand{\Cc}{\mathcal{C}}
\newcommand{\Dc}{\mathcal{D}}
\newcommand{\Ec}{\mathcal{E}}
\newcommand{\Fc}{\mathcal{F}}
\newcommand{\Gc}{\mathcal{G}}
\newcommand{\Hc}{\mathcal{H}}
\newcommand{\Ic}{\mathcal{I}}
\newcommand{\Jc}{\mathcal{J}}
\newcommand{\Kc}{\mathcal{K}}
\newcommand{\Lc}{\mathcal{L}}
\newcommand{\Mc}{\mathcal{M}}
\newcommand{\Nc}{\mathcal{N}}
\newcommand{\Oc}{\mathcal{O}}
\newcommand{\Pc}{\mathcal{P}}
\newcommand{\Qc}{\mathcal{Q}}
\newcommand{\Rc}{\mathcal{R}}
\newcommand{\Sc}{\mathcal{S}}
\newcommand{\Tc}{\mathcal{T}}
\newcommand{\Uc}{\mathcal{U}}
\newcommand{\Vc}{\mathcal{V}}
\newcommand{\Wc}{\mathcal{W}}
\newcommand{\Xc}{\mathcal{X}}
\newcommand{\Yc}{\mathcal{Y}}
\newcommand{\Zc}{\mathcal{Z}}

\newcommand{\Ab}{\mathbb{A}}
\newcommand{\Bb}{\mathbb{B}}
\newcommand{\Cb}{\mathbb{C}}
\newcommand{\Db}{\mathbb{D}}
\newcommand{\Eb}{\mathbb{E}}
\newcommand{\Fb}{\mathbb{F}}
\newcommand{\Gb}{\mathbb{G}}
\newcommand{\Hb}{\mathbb{H}}
\newcommand{\Ib}{\mathbb{I}}
\newcommand{\Jb}{\mathbb{J}}
\newcommand{\Kb}{\mathbb{K}}
\newcommand{\Lb}{\mathbb{L}}
\newcommand{\Mb}{\mathbb{M}}
\newcommand{\Nb}{\mathbb{N}}
\newcommand{\Ob}{\mathbb{O}}
\newcommand{\Pb}{\mathbb{P}}
\newcommand{\Qb}{\mathbb{Q}}
\newcommand{\Rb}{\mathbb{R}}
\newcommand{\Sb}{\mathbb{S}}
\newcommand{\Tb}{\mathbb{T}}
\newcommand{\Ub}{\mathbb{U}}
\newcommand{\Vb}{\mathbb{V}}
\newcommand{\Wb}{\mathbb{W}}
\newcommand{\Xb}{\mathbb{X}}
\newcommand{\Yb}{\mathbb{Y}}
\newcommand{\Zb}{\mathbb{Z}}

\newcommand{\av}{\mathbf{a}}
\newcommand{\bv}{\mathbf{b}}
\newcommand{\cv}{\mathbf{c}}
\newcommand{\dv}{\mathbf{d}}
\newcommand{\ev}{\mathbf{e}}
\newcommand{\fv}{\mathbf{f}}
\newcommand{\gv}{\mathbf{g}}
\newcommand{\hv}{\mathbf{h}}
\newcommand{\iv}{\mathbf{i}}
\newcommand{\jv}{\mathbf{j}}
\newcommand{\kv}{\mathbf{k}}
\newcommand{\lv}{\mathbf{l}}
\newcommand{\mv}{\mathbf{m}}
\newcommand{\nv}{\mathbf{n}}
\newcommand{\ov}{\mathbf{o}}
\newcommand{\pv}{\mathbf{p}}
\newcommand{\qv}{\mathbf{q}}
\newcommand{\rv}{\mathbf{r}}
\newcommand{\sv}{\mathbf{s}}
\newcommand{\tv}{\mathbf{t}}
\newcommand{\uv}{\mathbf{u}}
\newcommand{\vv}{\mathbf{v}}
\newcommand{\wv}{\mathbf{w}}
\newcommand{\xv}{\mathbf{x}}
\newcommand{\yv}{\mathbf{y}}
\newcommand{\zv}{\mathbf{z}}

\newcommand{\Av}{\mathbf{A}}
\newcommand{\Bv}{\mathbf{B}}
\newcommand{\Cv}{\mathbf{C}}
\newcommand{\Dv}{\mathbf{D}}
\newcommand{\Ev}{\mathbf{E}}
\newcommand{\Fv}{\mathbf{F}}
\newcommand{\Gv}{\mathbf{G}}
\newcommand{\Hv}{\mathbf{H}}
\newcommand{\Iv}{\mathbf{I}}
\newcommand{\Jv}{\mathbf{J}}
\newcommand{\Kv}{\mathbf{K}}
\newcommand{\Lv}{\mathbf{L}}
\newcommand{\Mv}{\mathbf{M}}
\newcommand{\Nv}{\mathbf{N}}
\newcommand{\Ov}{\mathbf{O}}
\newcommand{\Pv}{\mathbf{P}}
\newcommand{\Qv}{\mathbf{Q}}
\newcommand{\Rv}{\mathbf{R}}
\newcommand{\Sv}{\mathbf{S}}
\newcommand{\Tv}{\mathbf{T}}
\newcommand{\Uv}{\mathbf{U}}
\newcommand{\Vv}{\mathbf{V}}
\newcommand{\Wv}{\mathbf{W}}
\newcommand{\Xv}{\mathbf{X}}
\newcommand{\Yv}{\mathbf{Y}}
\newcommand{\Zv}{\mathbf{Z}}

\newcommand{\alphav     }{\boldsymbol \alpha     }
\newcommand{\betav      }{\boldsymbol \beta      }
\newcommand{\gammav     }{\boldsymbol \gamma     }
\newcommand{\deltav     }{\boldsymbol \delta     }
\newcommand{\epsilonv   }{\boldsymbol \epsilon   }
\newcommand{\varepsilonv}{\boldsymbol \varepsilon}
\newcommand{\zetav      }{\boldsymbol \zeta      }
\newcommand{\etav       }{\boldsymbol \eta       }
\newcommand{\thetav     }{\boldsymbol \theta     }
\newcommand{\varthetav  }{\boldsymbol \vartheta  }
\newcommand{\iotav      }{\boldsymbol \iota      }
\newcommand{\kappav     }{\boldsymbol \kappa     }
\newcommand{\varkappav  }{\boldsymbol \varkappa  }
\newcommand{\lambdav    }{\boldsymbol \lambda    }
\newcommand{\muv        }{\boldsymbol \mu        }
\newcommand{\nuv        }{\boldsymbol \nu        }
\newcommand{\xiv        }{\boldsymbol \xi        }
\newcommand{\omicronv   }{\boldsymbol \omicron   }
\newcommand{\piv        }{\boldsymbol \pi        }
\newcommand{\varpiv     }{\boldsymbol \varpi     }
\newcommand{\rhov       }{\boldsymbol \rho       }
\newcommand{\varrhov    }{\boldsymbol \varrho    }
\newcommand{\sigmav     }{\boldsymbol \sigma     }
\newcommand{\varsigmav  }{\boldsymbol \varsigma  }
\newcommand{\tauv       }{\boldsymbol \tau       }
\newcommand{\upsilonv   }{\boldsymbol \upsilon   }
\newcommand{\phiv       }{\boldsymbol \phi       }
\newcommand{\varphiv    }{\boldsymbol \varphi    }
\newcommand{\chiv       }{\boldsymbol \chi       }
\newcommand{\psiv       }{\boldsymbol \psi       }
\newcommand{\omegav     }{\boldsymbol \omega     }

\newcommand{\Gammav     }{\boldsymbol \Gamma     }
\newcommand{\Deltav     }{\boldsymbol \Delta     }
\newcommand{\Thetav     }{\boldsymbol \Theta     }
\newcommand{\Lambdav    }{\boldsymbol \Lambda    }
\newcommand{\Xiv        }{\boldsymbol \Xi        }
\newcommand{\Piv        }{\boldsymbol \Pi        }
\newcommand{\Sigmav     }{\boldsymbol \Sigma     }
\newcommand{\Upsilonv   }{\boldsymbol \Upsilon   }
\newcommand{\Phiv       }{\boldsymbol \Phi       }
\newcommand{\Psiv       }{\boldsymbol \Psi       }
\newcommand{\Omegav     }{\boldsymbol \Omega     }

%% file: 00_abstract.tex
\begin{abstract}

State estimation is a crucial component for the successful implementation of robotic systems, relying on sensors such as cameras, LiDAR, and IMUs. However, in real-world scenarios, the performance of these sensors is degraded by challenging environments, e.g. adverse weather conditions and low-light scenarios. The emerging 4D imaging radar technology is capable of providing robust perception in adverse conditions. Despite its potential, challenges remain for indoor settings where noisy radar data does not present clear geometric features. Moreover, disparities in radar data resolution and field of view (FOV) can lead to inaccurate measurements. While prior research has explored radar-inertial odometry based on Doppler velocity information, challenges remain for the estimation of 3D motion because of the discrepancy in the FOV and resolution of the radar sensor. In this paper, we address Doppler velocity measurement uncertainties. We present a method to optimize body frame velocity while managing Doppler velocity uncertainty. Based on our observations, we propose a dual imaging radar configuration to mitigate the challenge of discrepancy in radar data. To attain high-precision 3D state estimation, we introduce a strategy that seamlessly integrates radar data with a consumer-grade IMU sensor using fixed-lag smoothing optimization. Finally, we evaluate our approach using real-world 3D motion data.
\end{abstract}


%% file: 01_introduction.tex
\section{Introduction}

State estimation serves as a fundamental component in the majority of robotics applications. Commonly used sensors for state estimation include cameras, LiDAR, and IMUs \cite{zhao2021super}. However, deploying robots can be challenging in adverse environments. For instance, LiDAR sensors can experience substantial performance degradation when exposed to adverse factors such as smoke and fog, as well as in environments lacking distinct geometric features. Similarly, cameras encounter the same challenges with additional difficulties in environments with low-light conditions or lacking distinct visual features.

The emerging 4D mmWave imaging radar sensor \cite{iovescu2017fundamentals} can provide robust perception in demanding environments. The imaging radar sensor utilizes electromagnetic waves with wavelengths at the millimeter level that can function in adverse environmental conditions and do not rely on external lighting. Using frequency-modulated continuous wave technology, the imaging radar can provide Doppler velocity measurements for each detected 3D point. 
However, radar point clouds are known for being noisy and sparse and can be severely discretized due to post-processing procedures, thereby providing limited geometry information, especially in confined indoor environments.

Despite the challenges presented by the mmWave imaging radar, several recent
studies have found success using this technology for object detection \cite{wang2021rodnet, wang2021rod2021}, navigation \cite{huang2021cross} and state estimation \cite{park20213d, doer2021x, kramer2022coloradar, lu2020milliego}. Further applications are discussed in the survey paper \cite{venon2022millimeter}.

\begin{figure}[t]
    \centering

    \includegraphics[width=\linewidth]{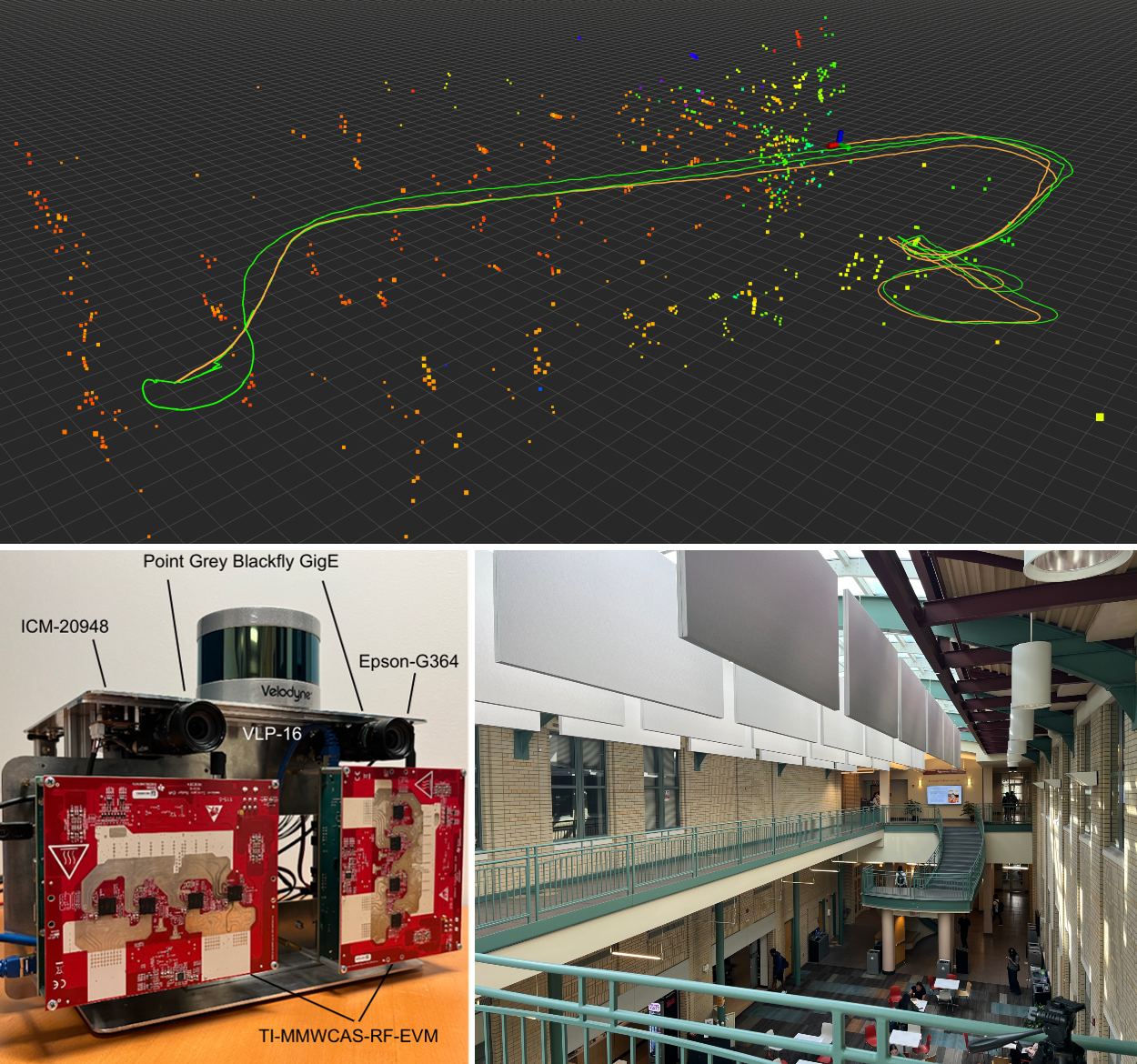}

    \caption{A demonstration of multi-radar inertial odometry (orange) walked through different levels of an atrium compared to visual-inertial odometry (green). The colors of the radar point cloud indicate Doppler velocity from high (red) to low (purple).}
    
    \label{fig:teaser}

    \vspace{-0.5cm}
\end{figure}

There are several challenges in developing a state estimation algorithm using mmWave radar. Previous works demonstrated the capability of building trajectories based on geometry information and scan-to-scan matching in automotive settings \cite{kung2021normal, hong2020radarslam, cen2019radar}. However, these methods are limited to planar localization and often exhibit degraded performance in indoor environments where there are limited key point features and clear geometry patterns in addition to the noise introduced by multi-path reflections of the signals.

Another popular approach involves leveraging Doppler speed information to optimize body frame velocity and then fusing it with an inertial measurement sensor \cite{kramer2020radar, DoerMFI2020, park20213d}. However, inaccuracies in body frame velocity and its associated uncertainty can lead to trajectory drift along the elevation direction. This issue arises because the antenna of imaging radar is designed to provide significantly lower data resolution and FOV in the elevation dimension compared to the azimuth dimension \cite{texas2020imaging}.


To achieve high-precision 3D motion state estimation using a Doppler velocity-based radar-inertial odometry approach, we tackle the above-mentioned challenges. We analyze radar measurement uncertainty and develop an algorithm for seamless radar-IMU fusion. We showcase the effectiveness of this approach by using dual mmWave cascade imaging radars coupled with a consumer-grade IMU sensor to obtain accurate 3D state estimates. Our contributions can be summarized as follows:
\begin{itemize}
    \item \textbf{Doppler and Velocity Uncertainty:} We assess the uncertainty of radar Doppler measurements and the resulting uncertainty on estimated linear body frame velocity for fusion with IMU sensor data.


    \item \textbf{Multi-Radar Inertial Odometry:} We employ a fixed-lag smoother optimization strategy capable of fusing IMU and multiple radar data to compensate for measurement uncertainty from each other and ensure robustness to outlier measurements.

    \item \textbf{Evaluate with 3D Motion:} We evaluate our method with hand-held collected data while traversing through different levels of indoor buildings.
    
\end{itemize}


%% file: 02_related_work.tex
\section{Related Work}

\subsection{Radar-Inertial Odometry}
Doppler velocity information from mmWave radar point clouds has been used to estimate trajectory in previous research. Kramer et al.~\cite{kramer2020radar} proposed a sliding window optimization approach with a robust loss kernel to fuse IMU and radar Doppler velocity measurements. They demonstrate the estimated body frame velocity is close to a Visual-Inertial solution. Doer et al. \cite{DoerMFI2020} separately estimated the body frame velocity from radar points using a RANSAC method to remove outlier measurements introduced by noise. Then they proposed an EKF-based approach to fuse radar body frame velocity with IMU and barometer measurements to estimate drone poses in indoor environments. However, their results show difficulty on estimating height correctly without the assistance of barometer data. 

Based on the previous EKF-based approach, Zhuang et al.\ further integrated the full radar SLAM system: 4D iRIOM \cite{zhuang20234d}. The authors proposed using a Graduated Non-Convexity method to remove outlier Doppler velocity measurements and estimate velocity from radar. They also introduce scan-to-map matching using radar points in the system to improve trajectory estimation. The loop closure was done using Scan Context \cite{kim2018scan} to detect similar geometry on the radar points map. They demonstrate their methods on a small ground robot which is limited to a planar environment.

Recent research by Michalczyk et al.~\cite{michalczyk2022tightly, michalczyk2023multi} proposed to build 3D landmarks from radar point clouds and fuse them with IMU using multi-state EKF. However, this method required a controlled environment where persistent landmarks with high radar cross sections are required. 

A learning-based approach \cite{lu2020milliego} by Lu et al.\ was proposed to fuse IMU and radar measurements to estimate trajectory. The authors use convolutional neural networks to build feature encoders for radar and recurrent neural networks to process IMU and the fused feature embeddings. However, their method was compared in \cite{park20213d} and found to have difficulty generalizing to new environments.


\subsection{Multi-Radar for State Estimation}

To further improve state estimation using mmWave radar sensor. Adding multiple radar sensors in an attempt to increase the FOV has appeared in some research. Doer et al.\ based on their previous work EKF-RIO \cite{DoerMFI2020} proposed the x-RIO \cite{doer2021x} using a triple radar setting to increase the horizontal FOV of the robot. The setting enables them to use the Manhattan World Assumption \cite{coughlan2000manhattan} on the radar point clouds to improve the trajectory estimation. In \cite{ng2021continuous} Ng et al.\ use a sliding window optimization approach to fuse 4 radar measurements with an IMU to jointly estimate body frame velocity and odometry in an autonomous car application. However, both of the above multi-radar methods were applied only in the planar setting and did not address the elevation drift problem caused by relatively lower resolution on the elevation dimension. 

Park et al.~\cite{park20213d} proposed a solution to improve 3D ego-motion estimation by adding a second ground-facing radar that operates in synchronization with the primary radar. They only extract 2D velocity on the high-resolution azimuth dimension from each radar to fuse with IMU. They proposed a radar velocity factor that fuses radar velocity with the IMU gyroscope. The authors evaluated their method by traversing through different levels of a construction site to prove their performance with 3D motion. In comparison with their method, we present a solution for fusing multiple radars and IMU data without discarding high-frequency accelerometer measurements and without the need to synchronously trigger the radars.

The publicly available dataset for radar state estimation Coloradar dataset \cite{kramer2022coloradar} provides a cascade imaging radar and a single-chip imaging radar both placed horizontally. The dataset was collected with mostly planar motion. Therefore, it's necessary to design our own radar sensors rig in order to collect 3D motion radar data for the research of radar-inertial state estimation.

%% file: 04_methodology.tex
\vspace{-1mm}
\section{Methodology}

    In this section, we introduce our approach to estimating trajectories using a Radar-Inertial system. Firstly, for each radar in the system, we estimate the linear velocity in the sensor frame and its associated uncertainty using radar Doppler information. Subsequently, the estimated linear velocity is constructed as a keyframe to be fused with the IMU accelerometer and gyroscope data, serving as a measurement constraint in the integration process. The complete pose is integrated and optimized in a sliding window using the aforementioned measurements. The full states to be optimized in each radar keyframe are as follows:
    \begin{align}
        \mathcal{X}_i = \Big[\Rv_i, \mathbf{p}_i, \vv_i, \bv^a_i, \bv^g_i \Big] \in \mathbb{R}^{15}
    \end{align}


\subsection{Velocity Estimation using Radar Doppler Velocity}
    One special property of the Imaging Radar is the Doppler velocity of the detected points. The physical significance of the Doppler velocity is the projection of the sensor's linear velocity in the direction of the point. Assuming that most of the points in one frame are static objects, we can establish the relationship between the Doppler velocity $\mathbf{d}_n \in \Rb^1$ of the point $n$, its 3D position $\rv_n \in \Rb^3$, and the sensor's linear velocity $\vv^s_j \in \Rb^3$ in frame $j$ with the following equation:
    \begin{align}
        -\mathbf{d}_n = (\vv^s_j)^\top \frac{\rv_n}{||\rv_n||}
    \end{align}
    To obtain an accurate linear velocity of the sensor that aligns with the most measured Doppler velocities, we can use a least-squares approach to solve for $\vv^s$. The noise of Doppler measurement can be modeled by a Gaussian distribution with variance $\Sigma_d\in \Rb^1$. The optimized sensor velocity will be:
    \begin{equation}
        \vv^s = \argmin_{\vv^s}  \Big( \sum_n || (\vv^s)^\top \frac{\rv_n}{||\rv_n||} + \mathbf{d}_n ||^2_{\Sigma_d}\Big)
        \label{eq:r_dop}
    \end{equation}
    We can find the marginalized covariance $\Sigma_{\vv^s}\in \Rb^{3\times3}$ on variable $\vv^s$ by inverting the information matrix in the system. 
    \begin{align}
        \Sigma_{\vv^s} = \left( A^\top\Sigma_d^{-1} A \right)^{-1}
        \label{eq:marginalized_cov}
    \end{align}
    Where $A\in\Rb^{n\times3}$ and each row of $A$ is a unit vector directing to the radar point $\rv_n/||\rv_n||$. Notice that due to the radar antenna design, the radar points distribution is uneven in azimuth dimension and elevation dimension. As a result, the estimated velocity uncertainty exhibits variations along the XYZ axes. Further details regarding the evaluation of Doppler measurement uncertainty $\Sigma_d$ and the estimated velocity uncertainty will be discussed in \ref{chapter:doppler_cov}.
    
    In reality, radar data often includes noisy measurements and non-stationary objects. To remove these outliers, previous works have explored the RANSAC approach \cite{doer2021x}, non-linear optimization with Cauchy robust loss \cite{kramer2020radar}, and a Graduated Non-Convexity (GNC) method \cite{zhuang20234d}. While RANSAC is non-deterministic, GNC requires extra iterations to adjust the kernel, and the Cauchy robust loss is sensitive to the initial value setting. Considering computation efficiency and accuracy, we use a Cauchy robust loss kernel to remove outliers and use the Levenberg–Marquardt optimizer to solve the system. The initial value of the variable $\vv^{s}$ was set using the preintegrated body frame velocity from IMU measurements and rotated to the sensor's coordinate. Please refer to the next section for IMU integration. 
    
    Finally, to fuse the sensor velocity from different radars with IMU. We treat the IMU frame as the body frame. The linear velocity and the corresponding covariance on the body frame are as follows:
    \begin{align}
        \vv^b &= \Rv_{r}\vv^s \\
        \Sigma_{\vv^b} &= \Rv_{r} \Sigma_{\vv^s} \Rv_{r}^\top
        \label{eq:body_cov}
    \end{align}
    Where $\Rv_r \in SO(3)$ is the rotation from the radar sensor coordinate to the IMU coordinate.
    
\subsection{Preintegrated IMU Factor}\label{chapter:imu_factor}
    As introduced in \cite{forster2017imu}, we can efficiently fuse the IMU sensor with other low frame rate measurements using preintegration on manifold to avoid the relinearization procedure when the keyframes linearization point changes. With sensor measured linear acceleration $\Tilde{\mathbf{a}}_k$ and angular velocity $\Tilde{\mathbf{\omega}}_k$ between keyframe $i$ and the next keyframe $j$. The preintegrated measurement of relative position $\Delta \Tilde{\mathbf{p}}_{ij}$, orientation $\Delta \Tilde{\Rv}_{ij}$, and velocity $\Delta \Tilde{\vv}_{ij}$ are:
    \begin{align}
        \Delta\mathbf{\Tilde{R}}_{ij} &= \prod_{k=i}^{j-1} \textit{Exp}((\Tilde{\mathbf{\omega}}_k-\bv^g_i)\Delta t) \\  
        \Delta\Tilde{\vv}_{ij} &= \sum_{k=i}^{j-1} \Delta \mathbf{\Tilde{R}}_{ik}(\Tilde{\mathbf{a}}_k-\bv^a_i)\Delta t\\
        \Delta\Tilde{\mathbf{p}}_{ij} &= \sum_{k=i}^{j-1}\left(\Delta \Tilde{\vv}_{ik}\Delta t + \frac{1}{2}\mathbf{\Tilde{R}}_{ik}(\Tilde{\mathbf{a}}_k-\bv^a_i)\Delta t^2\right)
    \end{align}
    Where $\bv^a_i$ and $\bv^g_i$ are the slow varying linear acceleration bias and angular velocity bias. The measurement constraint of the IMU between $2$ keyframes has the residual in the following form.
    \begin{align}
        \rv_{\Delta\Rv_{ij}}&=\textit{Log}\left(\Delta\Tilde{\Rv}_{ij}^\top\left(\Rv_i^\top\Rv_j\right)\right) \\
        \rv_{\Delta\vv_{ij}}&=\Rv_i^\top\left(\vv_j-\left(\vv_i+\mathbf{g}\Delta t_{ij}\right)\right) -\Delta\Tilde{\vv}_{ij} \\
        \rv_{\Delta\mathbf{p}_{ij}}&=\Rv_i^\top \left( \mathbf{p}_j- \left( \mathbf{p}_i+\vv\Delta t_{ij}+\frac{1}{2}\mathbf{g}\Delta t_{ij}^2 \right) \right)-\Delta\Tilde{\mathbf{p}}_{ij} \\
        \rv_{\Delta\bv_{ij}^a} &= \bv^a_j - \bv^a_i \\ 
        \rv_{\Delta\bv_{ij}^g} &= \bv^g_j - \bv^g_i
    \end{align}
    The combined preintegrated IMU residual is written as:
    \begin{align}
        \rv_{\Delta\Iv_{ij}} = \left[\rv_{\Delta\Rv_{ij}}, \rv_{\Delta\vv_{ij}}, \rv_{\Delta\mathbf{p}_{ij}}, \rv_{\Delta\bv_{ij}^a}, \rv_{\Delta\bv_{ij}^g}\right] \in \Rb^{15}
    \end{align}
    With covariance $\Sigma_{\Delta\Iv_{ij}}\in\Rb^{15\times15}$ which takes into account the noise in the estimated bias used for integration, it also preserves the correlation between the bias uncertainty and the preintegrated measurements' uncertainty. Part of the covariance matrix related to the IMU bias variables is used to describe slow-varying bias evolution, the magnitude of which is proportional to the preintegration time $\Delta t_{ij}$. More details about IMU preintegration on manifold can be found in \cite{forster2017imu}.

    \begin{figure}
        \centering
        \includegraphics[width=\linewidth]{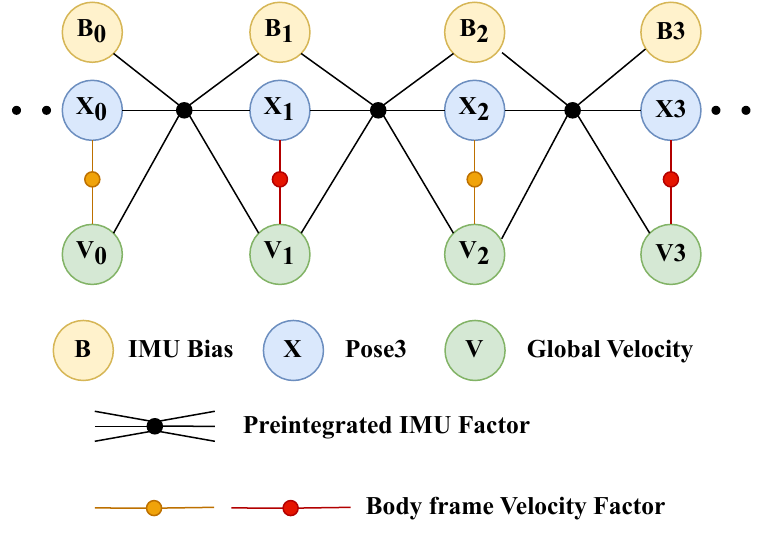}
        \vspace*{-6mm}
        \caption{An illustration of our multi-radar inertial state estimation system in the form of a factor graph. The body frame velocity factor can be built with either our horizontal or vertical imaging radar.}
        \label{fig:graph}
        \vspace{-7mm}
    \end{figure}

\subsection{Body Frame Velocity Factor}
    Upon receiving any body frame velocity $\vv^b_j$ estimated from the radar, we build a keyframe out of the preintegrated IMU measurements. This body frame velocity constraint will be added between the integrated global velocity variable $\vv_j$ and the rotation part of the pose variable $\Rv_j\in SO(3)$. By the definition of body frame velocity, the residual of this factor at frame $j$ is:
    \begin{align}
        \rv_{\Vv^b_j} = \Rv_j^\top\vv_j - \vv^b_j
        \label{eq:r_body}
    \end{align}

    Occasionally, inaccurate estimates of the body frame velocity lead to outlier residuals. To counteract this issue, we incorporate a Huber loss kernel on this body frame velocity residual when solving the system.

    To linearize the system, the Jacobian matrix with respect to rotation $\Jv_R$ and global velocity $\Jv_\vv$ are:
    \vspace{-1mm}
    \newcommand{\Skew}[1]{[#1]_{\times}}
    \begin{align}
        \Jv_R = \Skew{\Rv^\top\vv}\\
        \Jv_\vv = \Rv^\top
    \end{align}
    Where $\Skew{\Rv^\top\vv}$ is a skew symmetric matrix
    \begin{align}
        \Skew{\Rv^T\vv} = \Skew{\qv} = \begin{bmatrix}
            0 & -\qv_z & \qv_y \\ \qv_z & 0 & -\qv_x \\ -\qv_y & \qv_x & 0
        \end{bmatrix}
    \end{align}
    \vspace{-8mm}
    
\subsection{IMU Static Initialization}
    Since our preintegrated IMU factor is only constrained by the integrated velocity without a direct constraint within or among poses, an incorrect initial bias and rotation can easily cause the system to fail at the beginning. As IMU biases are modeled as slow-varying variables, it requires more steps to converge to the correct scale. Therefore, we have adapted and deployed a static initialization strategy for the IMU based on \cite{Geneva2020ICRA}, utilizing solely IMU measurements.

    The initialization process involves two consecutive sliding windows collecting IMU measurements. The second window, which includes the latest acceleration measurements, is used to detect any sensor movement by measuring the acceleration variance. Once motion is detected, we average the acceleration measurements in the first window to determine the gravity vector and its orthonormal basis using the Gram-Schmidt process. The initial rotation is determined by taking the inverse of the $SO(3)$ rotation formed by the orthonormal basis. The initial accelerometer bias is calculated by rotating the gravity constant using the initial rotation and then subtracting it from the measured gravity vector. The initial gyroscope bias is simply the mean of the angular velocities in the first measurement window. We have found this initialization procedure to be particularly crucial for uncalibrated IMUs with larger biases.
    
\subsection{Factor Graph Formulation}

    As IMU and radar measurements are inherently asynchronously triggered in time, IMU measurements were interpolated to be temporally aligned with radar measurements. This procedure guarantees that IMU measurements will always be available to construct a preintegrated IMU factor, even when two radar measurements are close and no IMU measurement exists in between.
    
    Given all the estimated measurements and their covariance. The full system optimizes states in a sliding window $\mathcal{S}$ and minimizes their Mahalanobis distance:
    \begin{equation}
        \begin{aligned}
        \mathcal{X}^*_\mathcal{S} &= \argmin_\mathcal{X_\mathcal{S}}\Bigg[ \sum_{i,j\in\mathcal{S}}\Big( ||\rv_{\Delta\Iv_{ij}}||^2_{\Sigma_\rv} + ||\rv_{\Vv^b_{j}}||^2_{\Sigma_{\vv^b}}\Big)\Bigg]
        \end{aligned}
    \end{equation}
    Where body frame velocity covariance $\Sigma_{\vv^b}$ was used for proper fusion with preintegrated IMU Factor.     
    
    An illustration of the system as a factor graph is shown in Fig.~\ref{fig:graph}. The aforementioned system was constructed using the \mbox{GTSAM} library \cite{gtsam}. The optimization problem is solved by a fixed-lag smoother with iSAM2 \cite{kaess2012isam2}. We set a 5-second optimization window where variables passed the window will be marginalized as prior factors.


%% file: 05_experiments.tex
\section{Experiments}

\subsection{Dual Cascade Imaging Radar Dataset}

We designed our handheld sensor platform to collect time-synchronized radar, IMU, and image data for evaluating multi-radar inertial state estimation. The platform includes a Velodyne VLP-16 LiDAR, two PointGrey Cameras running at $20\unit{\hertz}$, an Epson G-364 IMU running at $200\unit{\hertz}$, an ICM-20948 IMU running at $100\unit{\hertz}$, and two Texas Instrument Cascade Imaging Radars (MMWCAS-RF-EVM) running at $10\unit{\hertz}$. The sensor platform layout is depicted in Fig.~\ref{fig:teaser}. Timing is synchronized using an external Teensy MCU with pseudo GPS/PPS signals and trigger signals for the cameras.

The commercially available TI mmWave cascade imaging radar has a theoretical azimuth/elevation angular resolution of $1.4^\circ / 18^\circ$ \cite{texas2020imaging}. These resolutions represent the minimum angle between two equally large targets at the same range that the radar can distinguish and separate from each other. To compensate for this limitation, we employ two radars placed horizontally and vertically in this research. Our radar operates in Multi-Input Multi-Output (MIMO) mode. To avoid frequency conflicts, we set the start frequency of the horizontal and vertical radars as $77 \unit{\giga\hertz}$ and $79 \unit{\giga\hertz}$, with both radars having a chirp slope of $40 \unit{\mega\hertz\per\micro\second}$ and a bandwidth of approximately $1.5 \unit{\giga\hertz}$. With these radar signal settings, we achieve a maximum range of $30 \unit{\meter}$, a Doppler velocity resolution of $0.055 \unit{\meter\per\second}$, and a maximum Doppler velocity range of $\pm1.76 \unit{\meter\per\second}$. We employed a Constant False Alarm Rate (CFAR) algorithm provided by the TI mmWave studio to process the raw radar data into point clouds, including signal-to-noise ratio and Doppler velocity information for each point.

Unlike previous work \cite{park20213d}, where one radar was directed downward to the ground, we designed both radars to face forward. This decision was made to accommodate robots with limited space for installing downward-facing sensors. Additionally, our state estimation results suggest potential future enhancements in imaging radar designs to improve resolution in both azimuth and elevation dimensions.

We collected three sequences with mostly planar motion and three sequences involving 3D motion across various levels within the building. The data collection was conducted at a normal human walking pace and motion. The traverse length of each trajectory is detailed in TABLE \ref{table:eval}.

\subsection{Doppler and Velocity Uncertainty Evaluation}\label{chapter:doppler_cov}

To properly fuse all radar measurements with IMU measurements, we need to understand the noise levels of both sensors. The IMU measurement uncertainty can be found in manufacturers' manuals or determined by analyzing hours of static IMU data \cite{woodman2007introduction}. However, current research and the radar sensor manufacturer do not provide Doppler velocity measurement uncertainty, and the Doppler velocity resolution does not accurately represent it.

\begin{figure}
    \centering
    \includegraphics[width=\linewidth]{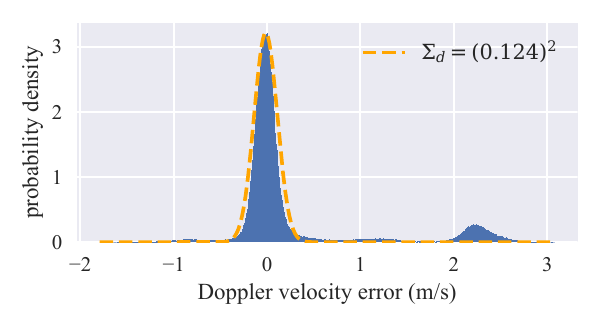}
    \vspace*{-9mm}
    \caption{The probability density of the Doppler velocity error distribution is calculated by comparing the velocities of 1,092,224 radar points with the VIO body-frame velocity projected in their respective directions.}
    \label{fig:doppler_err}
    \vspace{-7mm}
\end{figure}

To assess radar Doppler velocity errors, we utilize body-frame velocity data obtained from visual-inertial odometry pseudo ground truth. By projecting this velocity onto the direction of all radar points during a two-minute sequence(NSH\_atrium), we can analyze the Doppler velocity error distribution, as depicted in Fig.~\ref{fig:doppler_err}. The smaller peak in this bimodal distribution is the result of Doppler velocity values exceeding sensor limits. After excluding such outliers, we approximate the noise as a Gaussian distribution with variance $\Sigma_d \approx (0.124\ m/s)^2$. It's important to note that this uncertainty may vary with different radar signal settings, depending on factors such as Doppler velocity resolution and elevation/azimuth angle resolution.

\begin{figure}
    \centering
    \includegraphics[width=\linewidth]{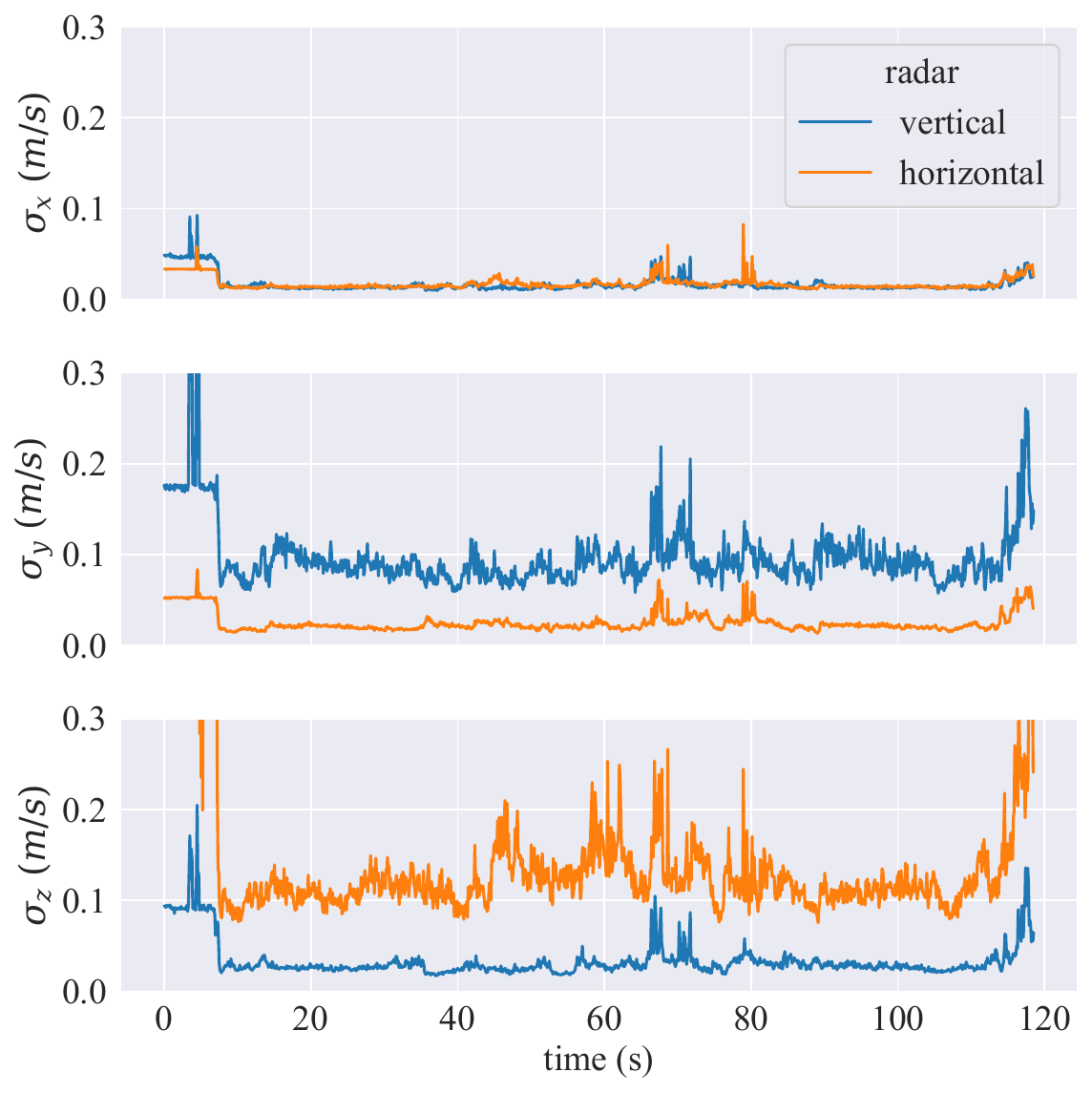}
    \vspace*{-7mm}
    \caption{Standard deviation on XYZ axes of the estimated body frame velocity from two radars.}
    \label{fig:radar_v_std}
    \vspace{-7mm}
\end{figure}

We use this Doppler velocity noise model to compute the marginalized covariance on the optimized sensor frame velocity, and then the corresponding uncertainty on body frame velocity follows equations (\ref{eq:marginalized_cov}) and (\ref{eq:body_cov}). Fig.~\ref{fig:radar_v_std} displays the standard deviation of the body frame linear velocity estimated from two different radars in a two-minute sequence. For each radar, the axis with high uncertainty aligns with the elevation direction. This emphasizes the importance of using multiple radars to compensate for inaccurate measurements from each other and to fuse them with the correct covariance scale. We also noticed that uncertainty increases when there are fewer radar measurements, such as at the sequence's start and end when sensor motion is nearly stationary, and around the 70-second mark when the radar faces an open area for a short period.

\subsection{Quantitative Evaluation}
\vspace{-1mm}
Given that our evaluation environment is indoors and includes long corridors, LiDAR point cloud registration becomes challenging in environments with such geometric features. Therefore we evaluate our method using pseudo ground truth generated by the stereo visual-inertial odometry system OpenVINS \cite{Geneva2020ICRA}, with two PointGrey cameras and the Epson IMU. Our radar-inertial system is running with ICM-20948 IMU. The final trajectory was assessed using the EVO library \cite{grupp2017evo}, without the use of any alignment algorithms. The results for absolute and relative pose errors are presented in TABLE \ref{table:eval}. From the results, it's apparent that using both radars leads to a slight decrease in rotation performance. However, we are able to significantly improve translational drift compared to using a single radar setting.

We had considered comparing our method with, to the best of our knowledge, the only open-sourced EKF-based radar-inertial odometry algorithm, X-RIO \cite{doer2021x}. However, their algorithm was primarily designed for single-chip mmWave radar and requires handcrafted radar velocity uncertainty, such as adjustments for offset and maximum limit of the radar velocity covariance, defining the threshold for rejecting radar velocity updates. We chose not to include this comparison in our evaluation, as we were unable to generate meaningful results.

\input{05_table}

\begin{figure*}
    \centering
    \includegraphics[width=\linewidth]{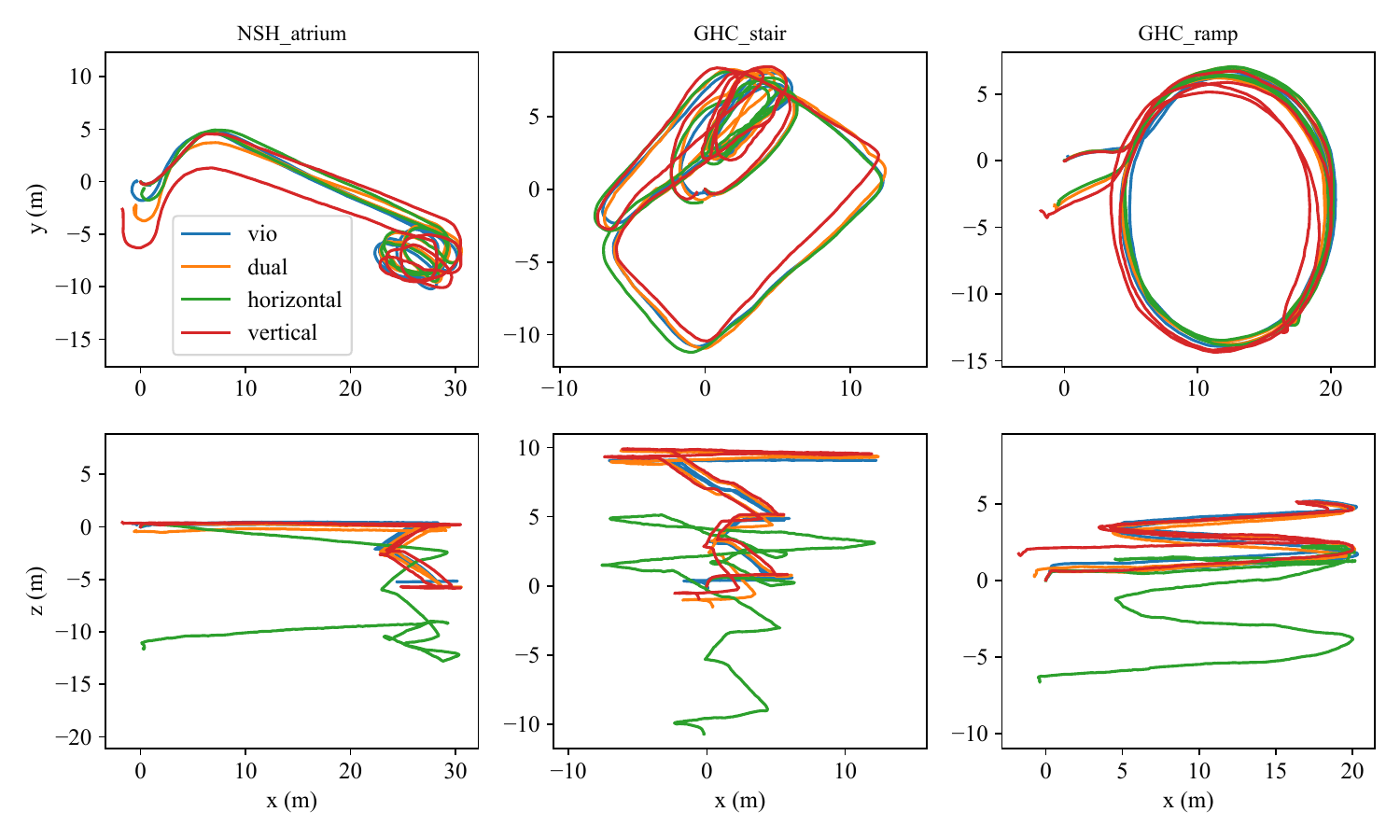}
    \vspace*{-9mm}
    \caption{Comparing the trajectories of visual-inertial odometry to radar-inertial odometry using either single horizontal/vertical radar or dual radars. Each column represents our 3D motion sequences, with the top row displaying the top-down view and the bottom row presenting the side view.}
    \label{fig:vis_odom}
    \vspace*{-7mm}
\end{figure*}

\subsection{Qualitative Evaluation}
\vspace{-1mm}
Fig.~\ref{fig:vis_odom} displays the estimated trajectory of our method compared to visual-inertial odometry. From the top view, it's clear that using only vertical radar results in increased translational drift compared to other settings, mainly due to its limited capacity to estimate lateral velocity. On the other hand, employing only a horizontal radar yields excellent performance in the XY plane but leads to substantial drift along the z-axis due to its reduced capacity to estimate vertical velocity effectively.

Given that human walking motion involves minimal lateral movement, the introduction of more vertical motion when traversing different levels of a building can result in significant drift when exclusively using horizontally placed radar sensors. This highlights the importance of considering motion primitives when deciding where to install radars for Doppler velocity-based radar-inertial odometry.

The attached video shows mapping results using estimated poses and radar points that have Doppler velocity projection error under the threshold. In the video, the geometry of the environment emerges even with noisy radar point clouds.



%% file: 05_table.tex
\begin{table}
\caption{RMSE of APE and RPE in 6 different sequences}
\centering
\resizebox{\columnwidth}{!}{%
\begin{tabular}{cccccc}
                 &                               & \multicolumn{2}{c}{APE}       & \multicolumn{2}{c}{RPE}          \\[-1.5ex] \\ \cline{3-6} & \\[-1.5ex]
sequence \& info                    & method     & trans         & rot           & trans         & rot              \\  \\[-1.5ex] \hline \\[-1.5ex]
\multicolumn{1}{c|}{2D}             & dual       & \textbf{1.48} & 2.73          & \textbf{0.08} & 0.66             \\
\multicolumn{1}{c|}{FRC\_1st}       & horizontal & 3.46          & \textbf{2.33} & 0.15          & \textbf{0.58}    \\
\multicolumn{1}{c|}{113 m}          & vertical   & 2.16          & 2.95          & 0.21          & 0.72             \\ \\[-1.5ex] \hline \\[-1.5ex]
\multicolumn{1}{c|}{2D}             & dual       & \textbf{1.31} & 3.76          & \textbf{0.09} & 0.65             \\
\multicolumn{1}{c|}{FRC\_2nd}       & horizontal & 2.88          & \textbf{3.40} & 0.12          & 0.69             \\
\multicolumn{1}{c|}{158 m}          & vertical   & 2.92          & 5.16          & 0.13          & \textbf{0.63}    \\ \\[-1.5ex] \hline \\[-1.5ex]
\multicolumn{1}{c|}{2D}             & dual       & \textbf{1.29} & 2.27          & \textbf{0.08} & 0.75             \\
\multicolumn{1}{c|}{NSH\_4th}       & horizontal & 4.07          & \textbf{1.52} & 0.12          & \textbf{0.66}    \\
\multicolumn{1}{c|}{167 m}          & vertical   & 1.77          & 2.74          & 0.12          & 0.68             \\ \\[-1.5ex] \hline  \\[-1.5ex]
\multicolumn{1}{c|}{3D}             & dual       & \textbf{1.09} & 4.92          & \textbf{0.10} & 0.89             \\
\multicolumn{1}{c|}{NSH\_atrium}    & horizontal & 7.22          & \textbf{2.33} & 0.18          & 0.88             \\
\multicolumn{1}{c|}{139 m}          & vertical   & 1.93          & 5.38          & 0.22          & \textbf{0.78}    \\ \\[-1.5ex] \hline  \\[-1.5ex]
\multicolumn{1}{c|}{3D}             & dual       & \textbf{1.16} & 8.13          & \textbf{0.11} & 0.63             \\
\multicolumn{1}{c|}{GHC\_stair}     & horizontal & 6.25          & \textbf{5.02} & 0.14          & \textbf{0.57}    \\
\multicolumn{1}{c|}{146 m}          & vertical   & 1.58          & 7.50          & 0.16          & 0.71             \\ \\[-1.5ex] \hline  \\[-1.5ex]
\multicolumn{1}{c|}{3D}             & dual       & \textbf{1.36} & 7.23          & \textbf{0.06} & \textbf{0.53}    \\
\multicolumn{1}{c|}{GHC\_ramp}      & horizontal & 4.09          & \textbf{6.43} & 0.09          & 0.60             \\
\multicolumn{1}{c|}{219 m}          & vertical   & 1.78          & 7.30          & 0.08          & \textbf{0.53}    \\ \\[-1.5ex] \hline  \\

\end{tabular}%
}

The units for APE: translation / rotation are meter and degree. The units for RPE: translation / rotation are percentage and degree per meter.

\label{table:eval}
\end{table}

%% file: 06_conclusion.tex
\vspace{-1mm}
\section{Conclusion}
\vspace{-1mm}
This paper presents a system using dual mmWave cascade imaging radars fused with an IMU sensor capable of achieving high-precision 3D motion state estimation. We provide insights into the limitations of radar measurements, emphasizing the resulting estimated uncertainty and the imperative need to compensate for inaccuracies through the use of multiple radars. We present our radar configurations and fixed-lag optimization solutions, which effectively integrate the radar and IMU measurements. Our method is demonstrated using a real-world 3D motion dataset. For further improvement, efforts can be made to explore the utilization of noisy radar point cloud geometry and to address issues related to point cloud registration degeneracy.

%% file: ICRA2024_MRIO.bbl
\begin{thebibliography}{10}
\providecommand{\url}[1]{#1}
\csname url@samestyle\endcsname
\providecommand{\newblock}{\relax}
\providecommand{\bibinfo}[2]{#2}
\providecommand{\BIBentrySTDinterwordspacing}{\spaceskip=0pt\relax}
\providecommand{\BIBentryALTinterwordstretchfactor}{4}
\providecommand{\BIBentryALTinterwordspacing}{\spaceskip=\fontdimen2\font plus
\BIBentryALTinterwordstretchfactor\fontdimen3\font minus
  \fontdimen4\font\relax}
\providecommand{\BIBforeignlanguage}[2]{{%
\expandafter\ifx\csname l@#1\endcsname\relax
\typeout{** WARNING: IEEEtran.bst: No hyphenation pattern has been}%
\typeout{** loaded for the language `#1'. Using the pattern for}%
\typeout{** the default language instead.}%
\else
\language=\csname l@#1\endcsname
\fi
#2}}
\providecommand{\BIBdecl}{\relax}
\BIBdecl

\bibitem{zhao2021super}
S.~Zhao, H.~Zhang, P.~Wang, L.~Nogueira, and S.~Scherer, ``Super odometry:
  {IMU}-centric {LiDAR}-visual-inertial estimator for challenging
  environments,'' in \emph{Proc. IEEE/RSJ Intl. Conf. on Intelligent Robots and
  Systems (IROS)}, Prague, {CZ}, Sep. 2021, pp. 8729--8736.

\bibitem{iovescu2017fundamentals}
C.~Iovescu and S.~Rao, ``The fundamentals of millimeter wave sensors,''
  \emph{Texas Instruments}, pp. 1--8, 2017.

\bibitem{wang2021rodnet}
Y.~Wang, Z.~Jiang, X.~Gao, J.-N. Hwang, G.~Xing, and H.~Liu, ``{RODNet}: Radar
  object detection using cross-modal supervision,'' in \emph{Proc. IEEE/CVF
  Winter Conference on Applications of Computer Vision (WACV)}, Waikoloa,
  {USA}, Jan. 2021, pp. 504--513.

\bibitem{wang2021rod2021}
Y.~Wang, J.-N. Hwang, G.~Wang, H.~Liu, K.-J. Kim, H.-M. Hsu, J.~Cai, H.~Zhang,
  Z.~Jiang, and R.~Gu, ``{ROD2021} challenge: A summary for radar object
  detection challenge for autonomous driving applications,'' in \emph{Proc. ACM
  Intl. Conf. on Multimedia Retrieval (ICMR)}, Taipei, {TW}, Nov. 2021, pp.
  553--559.

\bibitem{huang2021cross}
J.-T. Huang, C.-L. Lu, P.-K. Chang, C.-I. Huang, C.-C. Hsu, P.-J. Huang, H.-C.
  Wang \emph{et~al.}, ``Cross-modal contrastive learning of representations for
  navigation using lightweight, low-cost millimeter wave radar for adverse
  environmental conditions,'' \emph{IEEE Robotics and Automation Letters
  (RA-L)}, vol.~6, no.~2, pp. 3333--3340, 2021.

\bibitem{park20213d}
Y.~S. Park, Y.-S. Shin, J.~Kim, and A.~Kim, ``3{D} ego-motion estimation using
  low-cost mm{W}ave radars via radar velocity factor for pose-graph {SLAM},''
  \emph{IEEE Robotics and Automation Letters (RA-L)}, vol.~6, no.~4, pp.
  7691--7698, 2021.

\bibitem{doer2021x}
C.~Doer and G.~F. Trommer, ``{x-RIO}: Radar inertial odometry with multiple
  radar sensors and yaw aiding,'' \emph{Gyroscopy and Navigation}, vol.~12, pp.
  329--339, 02 2022.

\bibitem{kramer2022coloradar}
A.~Kramer, K.~Harlow, C.~Williams, and C.~Heckman, ``{ColoRadar}: The direct 3d
  millimeter wave radar dataset,'' \emph{Intl. J. of Robotics Research (IJRR)},
  vol.~41, no.~4, pp. 351--360, 2022.

\bibitem{lu2020milliego}
C.~X. Lu, M.~R.~U. Saputra, P.~Zhao, Y.~Almalioglu, P.~P. De~Gusmao, C.~Chen,
  K.~Sun, N.~Trigoni, and A.~Markham, ``{milliEgo}: single-chip mm{W}ave radar
  aided egomotion estimation via deep sensor fusion,'' in \emph{Proc. ACM Conf.
  on Embedded Networked Sensor Systems}, Yokohama, JP, Nov. 2020, pp. 109--122.

\bibitem{venon2022millimeter}
A.~Venon, Y.~Dupuis, P.~Vasseur, and P.~Merriaux, ``Millimeter wave {FMCW}
  radars for perception, recognition and localization in automotive
  applications: A survey,'' \emph{IEEE Trans. on Intelligent Vehicles (TIV)},
  vol.~7, no.~3, pp. 533--555, 2022.

\bibitem{kung2021normal}
P.-C. Kung, C.-C. Wang, and W.-C. Lin, ``A normal distribution transform-based
  radar odometry designed for scanning and automotive radars,'' in \emph{Proc.
  IEEE Intl. Conf. on Robotics and Automation (ICRA)}, Xi'an, {CN}, May 2021,
  pp. 14\,417--14\,423.

\bibitem{hong2020radarslam}
Z.~Hong, Y.~Petillot, and S.~Wang, ``Radar{SLAM}: Radar based large-scale
  {SLAM} in all weathers,'' in \emph{Proc. IEEE/RSJ Intl. Conf. on Intelligent
  Robots and Systems (IROS)}, Las Vegas, {USA}, Sep. 2020, pp. 5164--5170.

\bibitem{cen2019radar}
S.~H. Cen and P.~Newman, ``Radar-only ego-motion estimation in difficult
  settings via graph matching,'' in \emph{Proc. IEEE Intl. Conf. on Robotics
  and Automation (ICRA)}, Montreal, {CA}, 2019, pp. 298--304.

\bibitem{kramer2020radar}
A.~Kramer, C.~Stahoviak, A.~Santamaria-Navarro, A.-A. Agha-Mohammadi, and
  C.~Heckman, ``Radar-inertial ego-velocity estimation for visually degraded
  environments,'' in \emph{Proc. IEEE Intl. Conf. on Robotics and Automation
  (ICRA)}, Paris, {FR}, May 2020, pp. 5739--5746.

\bibitem{DoerMFI2020}
C.~Doer and G.~F. Trommer, ``An {EKF} based approach to radar inertial
  odometry,'' in \emph{Proc. Intl. Conf. on Multisensor Fusion and Integration
  for Intelligent Systems (MFI)}, Karlsruhe, {DE}, Sep. 2020, pp. 152--159.

\bibitem{texas2020imaging}
I.~Texas~Instruments, ``Imaging radar using cascaded mmwave sensor reference
  design,'' 2020.

\bibitem{zhuang20234d}
Y.~Zhuang, B.~Wang, J.~Huai, and M.~Li, ``4{D} {iRIOM}: 4{D} imaging radar
  inertial odometry and mapping,'' \emph{IEEE Robotics and Automation Letters
  (RA-L)}, vol.~8, no.~6, pp. 3246--3253, 2023.

\bibitem{kim2018scan}
G.~Kim and A.~Kim, ``Scan context: Egocentric spatial descriptor for place
  recognition within 3{D} point cloud map,'' in \emph{Proc. IEEE/RSJ Intl.
  Conf. on Intelligent Robots and Systems (IROS)}, Madrid, {ES}, Oct. 2018, pp.
  4802--4809.

\bibitem{michalczyk2022tightly}
J.~Michalczyk, R.~Jung, and S.~Weiss, ``Tightly-coupled {EKF}-based
  radar-inertial odometry,'' in \emph{Proc. IEEE/RSJ Intl. Conf. on Intelligent
  Robots and Systems (IROS)}, Kyoto, {JP}, Oct. 2022, pp. 12\,336--12\,343.

\bibitem{michalczyk2023multi}
J.~Michalczyk, R.~Jung, C.~Brommer, and S.~Weiss, ``Multi-state tightly-coupled
  {EKF}-based radar-inertial odometry with persistent landmarks,'' in
  \emph{Proc. IEEE Intl. Conf. on Robotics and Automation (ICRA)}, London,
  {UK}, May 2023, pp. 4011--4017.

\bibitem{coughlan2000manhattan}
J.~Coughlan and A.~L. Yuille, ``The {M}anhattan world assumption: Regularities
  in scene statistics which enable {B}ayesian inference,'' \emph{Advances in
  Neural Information Processing Systems}, vol.~13, 2000.

\bibitem{ng2021continuous}
Y.~Z. Ng, B.~Choi, R.~Tan, and L.~Heng, ``Continuous-time radar-inertial
  odometry for automotive radars,'' in \emph{Proc. IEEE/RSJ Intl. Conf. on
  Intelligent Robots and Systems (IROS)}, Prague, {CZ}, Sep. 2021, pp.
  323--330.

\bibitem{forster2017imu}
C.~Forster, L.~Carlone, F.~Dellaert, and D.~Scaramuzza, ``On-manifold
  preintegration for real-time visual--inertial odometry,'' \emph{IEEE Trans.
  on Robotics (TRO)}, vol.~33, no.~1, pp. 1--21, 2016.

\bibitem{Geneva2020ICRA}
P.~Geneva, K.~Eckenhoff, W.~Lee, Y.~Yang, and G.~Huang, ``{OpenVINS}: A
  research platform for visual-inertial estimation,'' in \emph{Proc. IEEE Intl.
  Conf. on Robotics and Automation (ICRA)}, Paris, {FR}, May 2020, pp.
  4666--4672.

\bibitem{gtsam}
\BIBentryALTinterwordspacing
F.~Dellaert and {GTSAM Contributors}, ``borglab/gtsam,'' May 2022. [Online].
  Available: \url{https://github.com/borglab/gtsam)}
\BIBentrySTDinterwordspacing

\bibitem{kaess2012isam2}
M.~Kaess, H.~Johannsson, R.~Roberts, V.~Ila, J.~J. Leonard, and F.~Dellaert,
  ``{iSAM2}: Incremental smoothing and mapping using the bayes tree,''
  \emph{Intl. J. of Robotics Research (IJRR)}, vol.~31, no.~2, pp. 216--235,
  2012.

\bibitem{woodman2007introduction}
O.~J. Woodman, ``An introduction to inertial navigation,'' University of
  Cambridge, Computer Laboratory, Tech. Rep., 2007.

\bibitem{grupp2017evo}
M.~Grupp, ``evo: Python package for the evaluation of odometry and {SLAM}.''
  \url{https://github.com/MichaelGrupp/evo}, 2017.

\end{thebibliography}
